\PassOptionsToPackage{dvipsnames,table,xcdraw}{xcolor}
\documentclass[10pt,twocolumn,letterpaper]{article}

\usepackage{iccv}              
\usepackage{multirow}
\usepackage{booktabs}
\usepackage{tabularx} 
\usepackage{array} 
\usepackage{amsmath} 
\usepackage{makecell} 

\usepackage{afterpage}

\definecolor{self_green}{HTML}{70AD47}
\definecolor{level_1}{HTML}{EDC1C0}
\definecolor{level_2}{HTML}{F6DFC1}
\definecolor{level_3}{HTML}{FEFBCF}

\definecolor{iccvblue}{rgb}{0.21,0.49,0.74}
\usepackage[pagebackref,breaklinks,colorlinks,allcolors=iccvblue]{hyperref}

\def\paperID{1274} 
\def\confName{ICCV}
\def\confYear{2025}

\title{TokenCarve: Information-Preserving Visual Token Compression \\ in Multimodal Large Language Models}

\author{
Xudong Tan$^{1}$\thanks{Equal Contribution} \quad Peng Ye$^{2,3 *}$ \quad Chongjun Tu$^{1}$ \quad Jianjian Cao$^{1}$ \quad Yaoxin Yang$^{1}$ \\ Lin Zhang$^{1}$\quad  Dongzhan Zhou$^{2}$ \quad Tao Chen$^{1}$\thanks{Corresponding Author: \textit{eetchen@fudan.edu.cn}} \\ 
$^{1}$Fudan University \quad $^{2}$Shanghai Artificial Intelligence Laboratory \\ 
\quad $^{3}$The Chinese University of Hong Kong \\
{\tt\small shawntan@126.com \quad eetchen@fudan.edu.cn}
}

\begin{document}
\maketitle
\begin{abstract}
Multimodal Large Language Models (MLLMs)  are becoming increasingly popular,
while the high computational cost associated with multimodal data input, particularly from visual tokens, poses a significant challenge. Existing training-based token compression methods improve inference efficiency but require costly retraining, while training-free methods struggle to maintain performance when aggressively reducing token counts. In this study, we reveal that the performance degradation of MLLM closely correlates with the accelerated loss of information in the attention output matrix. This insight introduces a novel information-preserving perspective, making it possible to maintain performance even under extreme token compression. Based on this finding, we propose TokenCarve, a training-free, plug-and-play, two-stage token compression framework. The first stage employs an Information-Preservation-Guided Selection (IPGS) strategy to prune low-information tokens, while the second stage further leverages IPGS to guide token merging, minimizing information loss. Extensive experiments on 11 datasets and 2 model variants demonstrate the effectiveness of TokenCarve. It can even reduce the number of visual tokens to 22.2\% of the original count, achieving a 1.23× speedup in inference, a 64\% reduction in KV cache storage, and only a 1.54\% drop in accuracy. Our code is available at \url{https://github.com/ShawnTan86/TokenCarve}.
\end{abstract}    
\section{Introduction}
\begin{figure}[t]
  \centering
      \includegraphics[width=\linewidth]{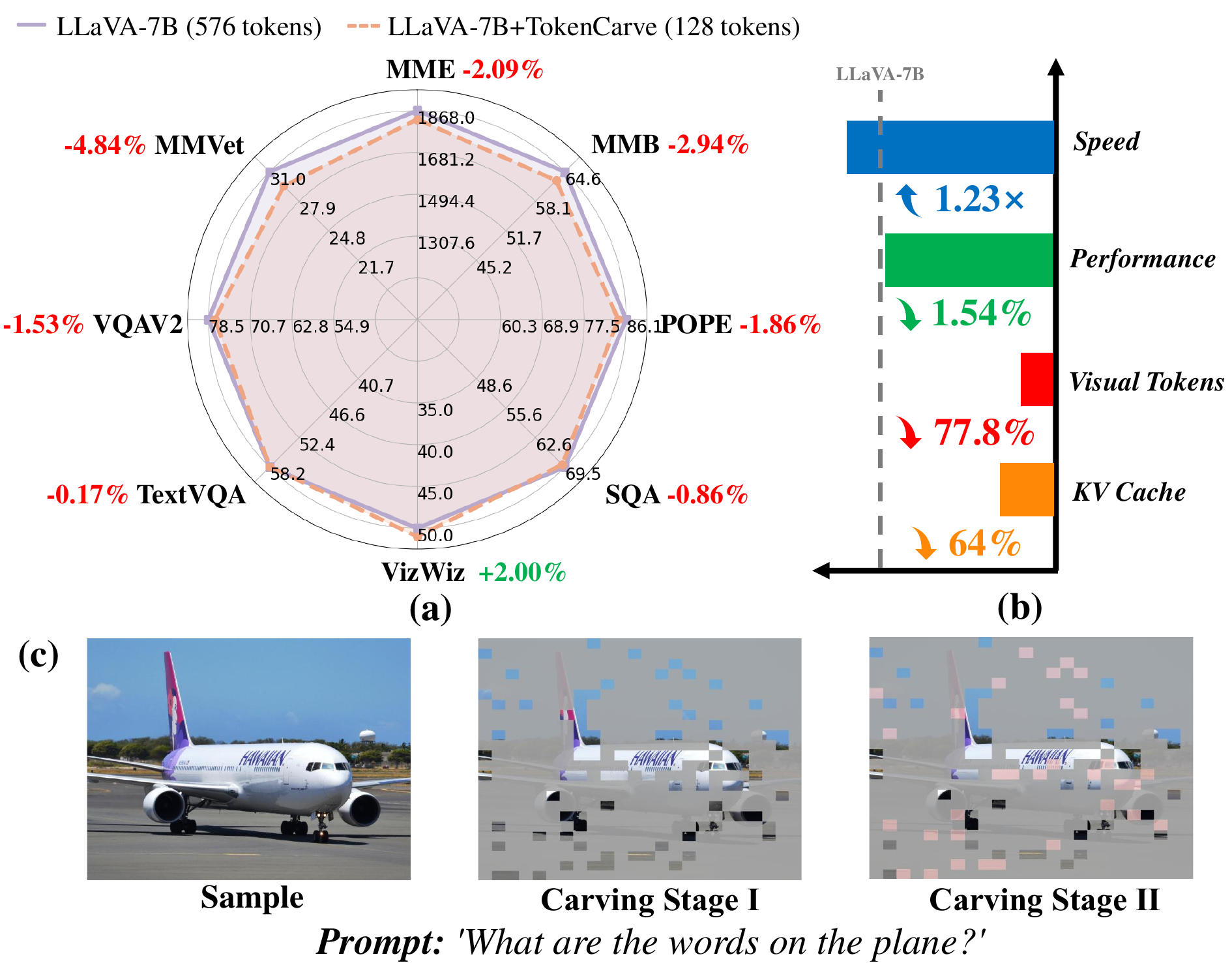}
\caption{(a) The radar chart illustrates the performance of TokenCarve on eight datasets when compressing the visual tokens of LLaVA1.5-7B from 576 to 192, showing that the overall performance remains close to the uncompressed version despite the significant token reduction; (b) Key performance indicators demonstrate that TokenCarve achieves a compression ratio of 77.8\%, with an average performance drop of only 1.54\%, a 1.23× inference speedup, and a 64\% reduction in KV cache usage; (c) The visualization depicts token positions during the two-stage compression process, where \textcolor[rgb]{0.65,0.65,0.65}{gray tokens} are pruned and \textcolor[rgb]{0.93,0.76,0.75}{pink tokens} are merged. In this OCR task example, TokenCarve consistently focuses on the critical regions of the image containing the text "Hawaii" throughout all compression stages.}
  \label{fig:1}  
  \vspace{-4mm}
\end{figure}

In recent years, the remarkable success of large language models~\cite{tokencarve_33_gpt4, tokencarve_34_claude, tokencarve_35_gemini} has not only fueled expectations for general artificial intelligence but also laid the foundation for the rapid advancement of Multimodal Large Language Models (MLLMs)~\cite{tokencarve_16_llava, tokencarve_37_deepseek, tokencarve_36_qwenvl, tokencarve_38_internvl}. By extending LLMs' powerful reasoning and generation capabilities to multimodal tasks such as image captioning, visual question answering, and cross-modal retrieval, MLLMs have significantly broadened the scope of AI applications. However, given that MLLMs typically contain billions—or even hundreds of billions—of parameters, achieving low-cost and efficient inference remains a critical challenge.

For most MLLMs, the computational overhead brought by visual tokens is substantial, leading to an \(\mathcal{O}(n^2)\) attention complexity and necessitating efficient compression strategies. Existing compression approaches for MLLMs can be broadly categorized into two types: training-based methods~\cite{tokencarve_27_llavamini, tokencarve_28_dynamic_llava, tokencarve_39_Training_MMLMs_1}, which rely on end-to-end learned compressors (e.g., query modules or Gumbel-softmax predictors) and thus require costly retraining and model-specific adaptation; and training-free methods, which include token compression applied to the visual encoder before feeding into the LLM~\cite{tokencarve_29_folder, tokencarve_31_llava_prumerge} as well as token compression within the internal layers of the LLM~\cite{tokencarve_11_sparsevlm, tokencarve_14_pyramiddrop}. While these training-free methods can effectively compress tokens,
their performance generally becomes difficult to maintain when the token count is significantly reduced.

In this study, we adopt a novel information-preservation perspective to investigate and design a training-free approach that sustains strong performance even under substantial token compression. Similar to~\cite{tokencarve_42_supporting_work_1, tokencarve_43_supporting_work_2, tokencarve_44_supporting_work_3, tokencarve_44_supporting_work_4}, we measure the information quantity of the token sequence based on the rank of the attention output matrix. Our experiments reveal a strong correlation between MLLM performance and information quantity. Specifically, we observe that as the token count decreases, both MLLM performance and information quantity exhibit a similar downward trajectory. Notably, when the number of visual tokens is reduced to approximately half, the rate of performance degradation intensifies, coinciding with a more rapid decline in information quantity and a discernible inflection point. This observation suggests that mitigating the rate of information loss during token compression may effectively postpone performance degradation in MLLMs.

Building on this insight, we introduce \textbf{TokenCarve}, an information-preserving, training-free, and plug-and-play visual token compression method. The core framework of TokenCarve employs a two-stage iterative token compression strategy. In the first stage, we innovatively propose an Information-Preservation-Guided Selection (\textbf{IPGS}) strategy, which leverages singular value decomposition on the outputs of the second visual token layer to quantify each token's contribution to the overall information quantity. By integrating these measurements with attention scores derived from the attention matrix, IPGS enables an initial pruning of tokens, striking a balance between information richness and cross-modal alignment. In the second stage, TokenCarve utilizes the token scores computed by IPGS to categorize the remaining tokens into two groups based on their relative importance. A fine-grained merging process is then applied, guided by token similarity, further reducing the token count while ensuring the retention of critical information. TokenCarve not only significantly reduces computational overhead but also effectively preserves key information, maintaining robust performance even when the token count is greatly reduced.

To evaluate the effectiveness of \textbf{TokenCarve}, we conduct comprehensive and systematic performance assessments across 11 widely used datasets, covering various tasks such as visual recognition, reasoning, OCR, and multimodal scientific question answering. Among these, results from 8 datasets are presented in the main result, while the remaining 3 are included in the supplementary materials. Moreover, we perform experiments on both 7B and 13B model variants. As illustrated in Figure~\ref{fig:1}, with the two-stage iterative token compression strategy, \textbf{TokenCarve} can preserve only 22.2\% of the original visual tokens while maintaining high accuracy, with merely a 1.54\% reduction in performance. Concurrently, the inference speed is enhanced by a factor of 1.23, and the KV cache storage is reduced by 64\%, leading to significant reductions in latency and improved storage efficiency. 

Our key contributions are summarized as follows:
\begin{enumerate}
    \item We reveal a correlation between MLLMs performance and the information quantity in the attention output matrix. Specifically, as the number of visual tokens decreases, both performance and information quantity exhibit a similar declining trend, with the decline accelerating when the token count reaches approximately half. This finding provides a novel information-preserving perspective for designing token compression methods.
    \item We propose \textbf{TokenCarve}, a two-stage token compression method. The first stage employs an Information Preservation Guided Selection (\textbf{IPGS}) strategy to select tokens with high information quantity. The second stage further uses IPGS to guide token merging, minimizing information loss. TokenCarve is training-free and plug-and-play, maintaining strong performance even when significantly reducing token count.
    \item We conduct extensive experiments to show the effectiveness of the proposed method. Specifically, \textbf{TokenCarve} effectively reduces the visual token count to 22.2\% of the original, achieving a 1.23× improvement in inference speed, a 64\% reduction in KV cache storage, and only a 1.54\% drop in accuracy. These results highlight its capability to substantially reduce computational overhead, mitigate latency, and optimize storage efficiency.
\end{enumerate}

\section{Related Work}

In multimodal large language models (MLLMs), visual tokens exhibit high information redundancy, leading to significant computational overhead~\cite{tokencarve_9_fastv, tokencarve_47_madtp, tokencarve_48_once}. Existing token compression methods for MLLMs can be broadly categorized into training-based methods~\cite{tokencarve_28_dynamic_llava,tokencarve_27_llavamini,tokencarve_39_Training_MMLMs_1,tokencarve_41_training1,tokencarve_41_training2,tokencarve_45_LLaVolta, tokencarve_46_deepstack} and training-free methods~\cite{tokencarve_9_fastv,tokencarve_11_sparsevlm,tokencarve_12_mustdrop,tokencarve_41_trainingfree1,tokencarve_41_trainingfree2, tokencarve_13_visionzip, tokencarve_31_llava_prumerge, tokencarve_29_folder,tokencarve_14_pyramiddrop}.

\textbf{Training-based visual token compression method for MLLMs.} QueCC~\cite{tokencarve_39_Training_MMLMs_1} explores a visual encoding structure in which the visual encoding stage requires only a single token input for vision token training. 
Dynamic-LLaVA~\cite{tokencarve_28_dynamic_llava} reveals that for visual reasoning tasks, optimal inference performance is achieved by using the largest LLM that fits within the computational budget while minimizing the number of visual tokens. LLaVA-Mini~\cite{tokencarve_27_llavamini} introduces the modality prefusion, which integrates visual information into text tokens beforehand, enabling extreme compression where only a single visual token is fed into the LLM backbone. However, these training-based methods require expensive retraining to adapt to different models, which limits their widespread adoption.

\textbf{Training-free visual token compression method for MLLMs.} FastV~\cite{tokencarve_9_fastv} identifies the phenomenon of inefficient attention in MLLMs and proposes that an image is worth 1/2 tokens after layer 2 of MLLMs. SparseVLM~\cite{tokencarve_11_sparsevlm} proposes a text-guided and training-free visual token optimization method that evaluates the importance of visual tokens by selecting text tokens that are the most relevant to visual content. MustDrop~\cite{tokencarve_12_mustdrop} designs dedicated visual token compression modules in the visual encoding, pre-filling, and decoding stages of MLLMs. VisionZip~\cite{tokencarve_13_visionzip} introduces a text-independent approach that significantly reduces the number of visual tokens by selecting dominant visual tokens and employing a similarity-based token merging strategy. Although these training-free methods mitigate visual token redundancy to some extent, they suffer from substantial performance degradation under high compression rates. To relieve this problem, we propose TokenCarve, a novel training-free visual token compression method that preserves essential information in MLLMs. By strategically retaining key visual tokens, TokenCarve reduces performance degradation and maintains strong performance even at extremely high compression rates.

\section{Methodology}
\subsection{Preliminaries}
MLLMs integrate both visual and textual inputs to enhance language understanding and generation. A typical architecture, exemplified by \textbf{LLaVA}~\cite{tokencarve_16_llava} and QwenVL~\cite{tokencarve_36_qwenvl}, comprises a visual encoder, a projection module, and an LLM. Given an image–text pair, the visual encoder (e.g., CLIP-ViT) extracts image features and converts them into \textbf{visual tokens} $\mathbf{T}_v$. These tokens are subsequently projected into the textual embedding space, thereby yielding contextualized visual embeddings. In parallel, the textual input is segmented into system tokens $\mathbf{T}_s$ and prompt tokens $\mathbf{T}_p$, where $\mathbf{T}_s$ encodes system-level instructions and $\mathbf{T}_p$ comprises user-provided prompts. The resulting multimodal sequence, $\mathbf{X} = [\mathbf{T}_s, \mathbf{T}_v, \mathbf{T}_p]$, is then input into the LLM to facilitate autoregressive token generation.

Within each Transformer layer, the model dynamically attends to both visual and textual tokens via a self-attention mechanism augmented with Rotary Position Embedding. The attention are computed as:
\begin{equation}
    \mathbf{A}^{l} = \text{softmax} \left( \frac{\mathbf{Q}^{l} R(\theta) (\mathbf{K}^{l} R(\theta))^T}{\sqrt{d_k}} \right),
    \label{eq:3-1}
\end{equation}
where $R(\theta)$ denotes a position-dependent rotation matrix applied to both the query $\mathbf{Q}^{l}$ and key $\mathbf{K}^{l}$ in the $l$-th layer, thereby ensuring that relative positional information is directly embedded within the attention mechanism~\cite{tokencarve_19_transformer}.

Subsequently, these attention scores serve to aggregate information from the value matrix:
\begin{equation}
    \mathbf{Z}^{l} = \mathbf{A}^{l} \mathbf{V}^{l},
    \label{eq:3-2}
\end{equation}
which is defined as the \textbf{attention output matrix} at layer $l$, encapsulating the contextualized representation of the multimodal sequence. For the final layer, $\mathbf{Z}^{l}$ is processed by a feedforward network to produce the ultimate hidden state $\mathbf{H}^{l}$, which is then projected via the language modeling head into the vocabulary space for autoregressive token generation~\cite{tokencarve_17_gpt3,tokencarve_18_llama}.

\subsection{Key Insight: Correlation Between MLLM Performance and the Information Quantity  of the Attention Output Matrix}
\label{key insight}
Taking LLaVA-7B as an example, during the prefilling stage the model employs 576 visual tokens out of a total of approximately 700 tokens, so that visual tokens constitute over 80\% of the input. Consequently, visual tokens represent the majority of the computational cost, and our objective is to reduce their number while minimizing performance degradation.

During our inference acceleration experiments that reduce the number of visual tokens, we observe that the model’s performance is intimately tied to the information quantity present in the Attention Output Matrix. This observation, in turn, motivates our subsequent experimental investigations.

\begin{figure}[!t]
\vspace{-4mm}
  \centering
  \includegraphics[width=\linewidth]{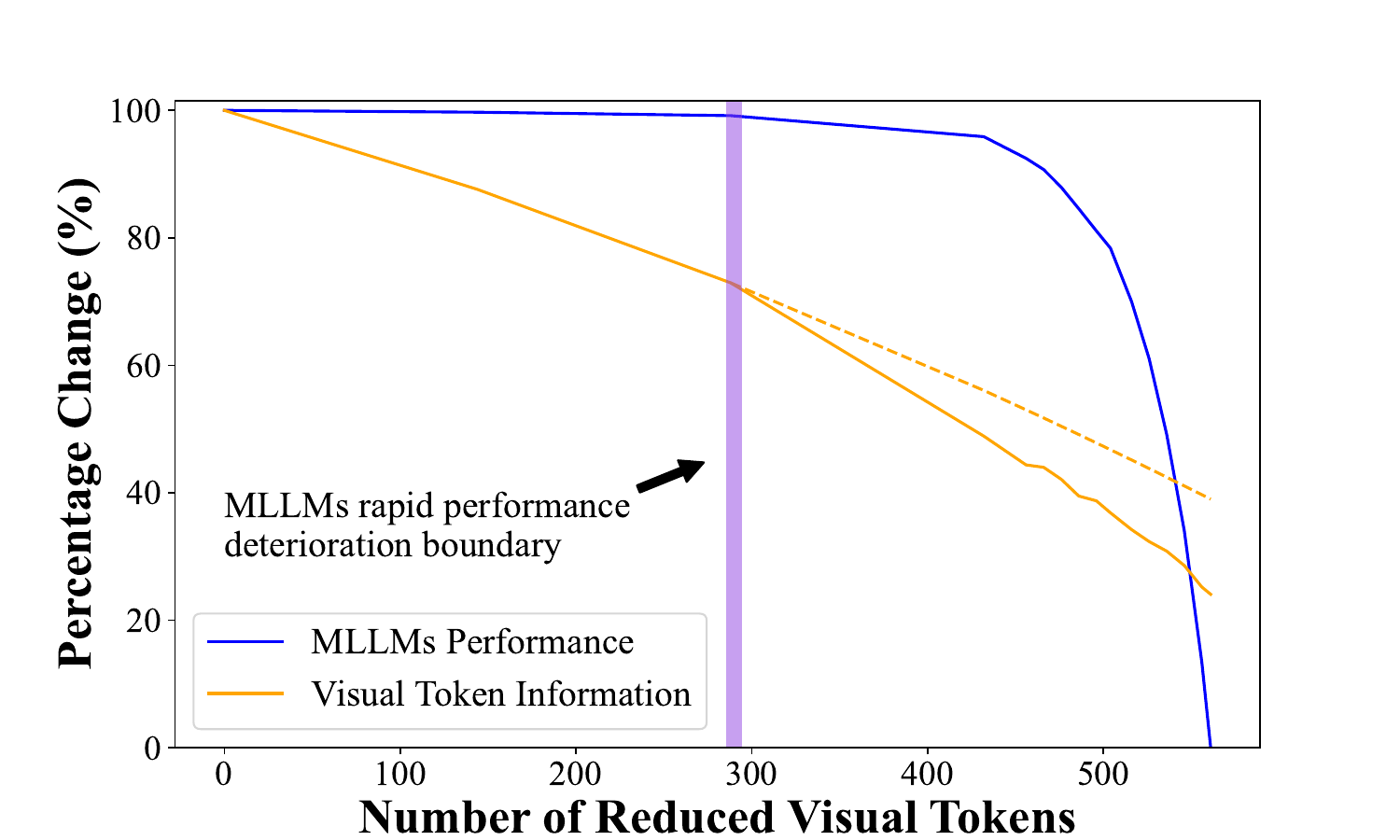}
\caption{\textbf{Key observation:} Curves of MLLM Performance and Visual Token Information under Visual Token Compression. The blue curve represents the performance variation, while the orange solid curve depicts the change in visual token information (with the orange dashed line indicating the original trend). The purple vertical line marks the threshold at which performance exhibits a pronounced decline, coinciding with an inflection in the information curve.}
  \label{fig:3-1}  
  \vspace{-4mm}
\end{figure}

\subsubsection{Visual Token Compression via Attention Score Selection}
We employ a token selection method based on attention scores to compress visual tokens on the OCRVQA~\cite{tokencarve_15_ocrvqa} dataset, progressively reducing their number from 576 to 15. By normalizing the derived performance curves using~\cref{eq:3-3}, we obtain the \textbf{MLLMs Performance} curve presented in~\cref{fig:3-1}.

\begin{equation}
    P_{\text{norm}} = \frac{P(v) - P(v_{min})}{P(v_{max}) - P(v_{min})},
    \label{eq:3-3}
\end{equation}
where $P(v)$ denotes the performance metric obtained with $v$ visual tokens, and $P(v_{\max})$ and $P(v_{\min})$ represent the maximum and minimum observed values, respectively.

\subsubsection{Analyzing the Information Quantity in the Attention Output Matrix}
During the prefilling stage, we extract the Attention Output Matrix from the final LLM layer ($Z^{L}$, for an LLM comprising $L$ layers) and isolate its visual component:
\[
Z_{\text{visual}}^{L} = Z^{L}[L_s: L_s + L_v],
\]
with $L_s$ and $L_v$ denoting the number of system tokens and visual tokens, respectively.

To quantify the information quantity of $Z_{\text{visual}}^{L}$, we compute its \textbf{matrix rank} as a proxy metric. By replicating the visual token compression process, we plot the matrix rank of \(Z_{\text{visual}}^{L}\) as the visual token count decreases from 576 to 15, and normalize this curve using~\cref{eq:3-3} to obtain the \textbf{Visual Token Information} curve shown in~\cref{fig:3-1}.

\begin{figure*}[h]
  \centering
  \includegraphics[width=\linewidth]{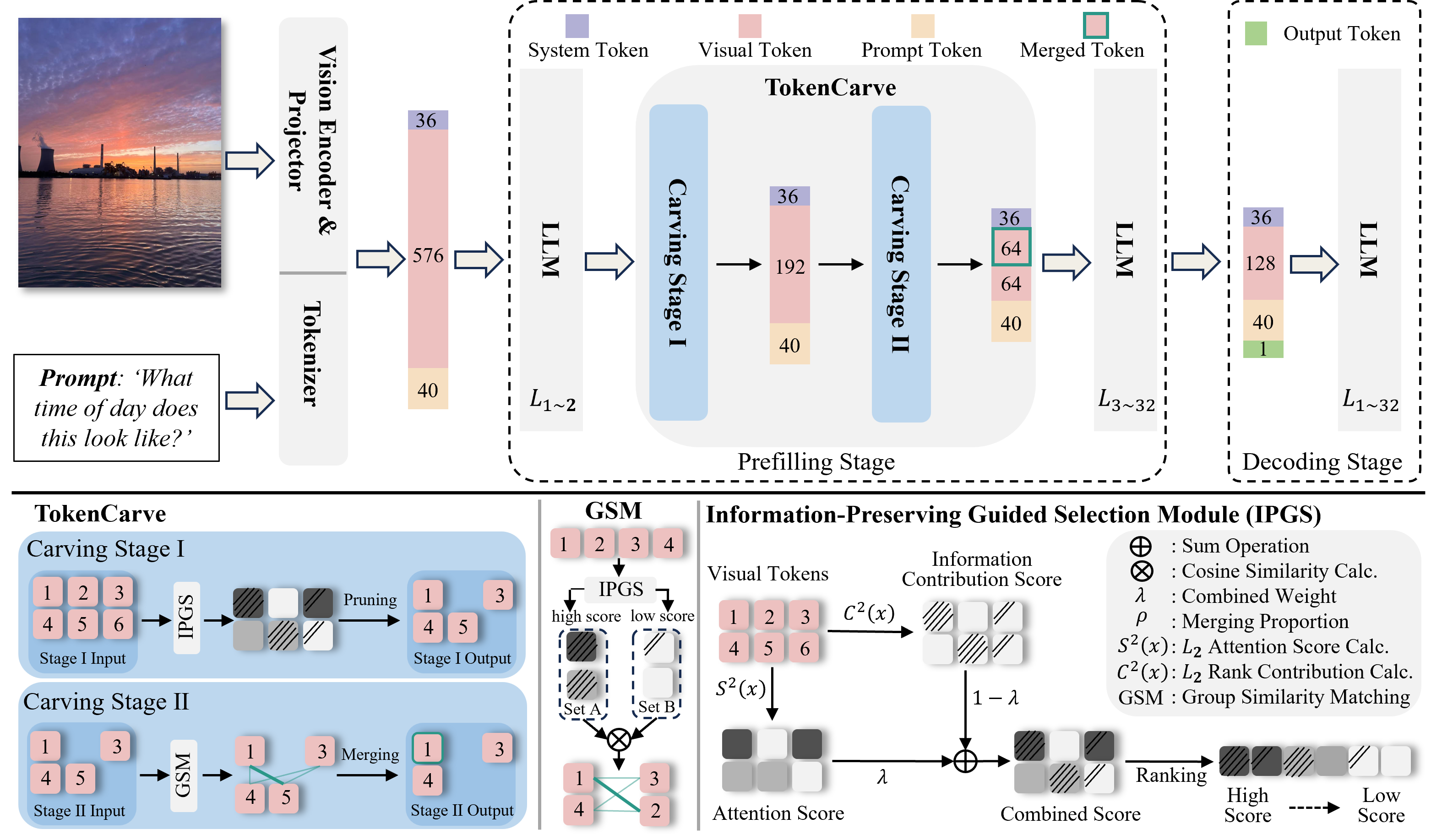}

\caption{The pipeline of the proposed TokenCarve framework. TokenCarve is integrated between the second and third layers of the LLaVA-1.5 model as a plug-and-play module, effectuating visual token compression during the prefilling stage. \textbf{\normalsize The upper panel} illustrates TokenCarve’s integration with LLaVA-1.5, which conventionally comprises 36 system tokens, 576 visual tokens, and a variable number of prompt tokens. \textbf{\normalsize The lower panel} (the blue region on the left) details the two-stage compression process: \textbf{Carving Stage I} employs the IPGS module to excise tokens with low contribution; \textbf{Carving Stage II} implements finer-grained token merging based on GSM, maximizing information retention.
\textbf{\normalsize The IPGS module} (right region of the lower panel) calculates each token’s information contribution score and attention score, then combines the two into a final ranking score (a higher slash count indicates a higher information contribution, and darker tokens signify higher attention). \textbf{\normalsize The GSM module} (middle region of the lower panel) uses these IPGS scores to split tokens into a higher-scored Set A and a lower-scored Set B, then merges Set B tokens with their most similar counterparts in Set A based on cosine similarity (with deeper connection lines representing higher similarity).}
\label{fig:3-2}
\vspace{-2mm}
\end{figure*}

\subsubsection{Observational Analysis}
From~\cref{fig:3-1}, we observe that early visual token compression has minimal impact on MLLM performance. However, once compression enters the purple-shaded region, performance declines in parallel with an accelerated loss of information in \(Z_{\text{visual}}^{L}\) due to its reduced rank. These results reveal a two-phase information loss process---initial gradual loss followed by rapid degradation---suggesting a close relationship between MLLM performance and the rank of the Attention Output Matrix.

This insight inspired our TokenCarve method, which adopts an \emph{Information-Preserving Guided Selection} approach to minimize information loss while optimizing visual token compression. By strategically preserving essential data, TokenCarve effectively delays performance degradation under extreme compression conditions, thereby enhancing the model’s overall efficiency.

\subsection{TokenCarve}

TokenCarve is inserted between the second and third layers of the LLM during the prefilling stage of MLLMs. It employs a two-stage approach to iteratively compress tokens, sculpting the original visual token sequence into a more compact form while minimizing information quantity loss. \cref{fig:3-2} illustrates the overall pipeline of TokenCarve, and this section provides a detailed explanation of the two carving stages.

\subsubsection{Carving Stage I: Using an Information-Preserving Guided Selection Strategy for Token Pruning}
Inspired by the observations in~\cref{key insight}, where the information quantity of MLLMs changes during inference, we propose the Information-Preserving Guided Selection (IPGS) strategy. This strategy identifies visual tokens that both contribute significantly to the overall information quantity and maintain high Attention Scores, ensuring an optimal balance between information richness and model preference.

To compute a token's contribution to the overall information quantity, we use the ~\cref{eq:3-3} obtained from the second LLM layer during the prefilling stage, denoted as $Z_{\text{visual}}^2 \in \mathbb{R}^{L_v \times d}$, where $L_v$ represents the number of visual tokens, and $d$ is the hidden dimension of the attention output matrix. By applying Singular Value Decomposition, $Z_{\text{visual}}^2$ can be expressed as $Z_{\text{visual}}^2 = U \Sigma V^T$, where $U$ is an $L_v \times L_v$ orthogonal matrix with column vectors as left singular vectors, $V$ is a $d \times d$ orthogonal matrix with column vectors as right singular vectors, and $\Sigma$ is an $L_v \times d$ diagonal matrix whose diagonal elements are singular values $\sigma_i$, typically sorted in descending order.

Each row of $Z_{\text{visual}}^2$ (corresponding to each token) can be represented as:
\begingroup
\setlength{\abovedisplayskip}{1pt}    
\setlength{\belowdisplayskip}{1pt}    
\begin{equation}
    Z_{\text{visual}}^2(x) = \sum_{i=1}^{r} u_{xi}\, \sigma_i\, v_i^T ,
    \label{eq:3-5}
\end{equation}
\endgroup
where $u_{xi}$ is the element at row $x$ and column $i$ in matrix $U$, indicating the projection coefficient of the $x$-th token onto the $i$-th left singular vector. Using this, the contribution of the $x$-th token to the rank of $Z_{\text{visual}}^2$ (i.e., its impact on the overall information quantity) can be computed as:
\begingroup
\setlength{\abovedisplayskip}{1pt}    
\setlength{\belowdisplayskip}{1pt}    
\begin{equation}
    C^2(x) = \sum_{i=1}^{r} \left| u_{xi}\, \sigma_i \right|.
    \label{eq:3-6}
\end{equation}
\vspace{-2mm}
\endgroup

For computing the \textbf{Attention Score of visual tokens}, we follow the approach in FastV, using the Attention Matrix from the second LLM layer, denoted as $A^2 \in \mathbb{R}^{H \times L_v \times L_v}$, where $H$ is the number of attention heads. The Attention Score for each visual token is obtained as follows:
\begingroup
\setlength{\abovedisplayskip}{1.5pt}    
\setlength{\belowdisplayskip}{1.5pt}    
\begin{equation}
    S^2(x) = \text{mean}\big(A^2[:,:,x]\big).
    \label{eq:3-7}
\end{equation}
\endgroup

By combining the \textbf{Information Contribution Score (ICS)} $C^2(x)$ and the \textbf{Attention Score (AS)} $S^2(x)$, we define the  combined score as:
\begingroup
\setlength{\abovedisplayskip}{1.5pt}    
\setlength{\belowdisplayskip}{1.5pt}    
\begin{equation}
    E^2(x) = (1-\lambda) \, C^2(x) + \lambda \, S^2(x),
    \label{eq:3-8}
\end{equation}
\endgroup
where $\lambda \in [0,1]$ is the weighting coefficient that balances the two metrics. Using $E^2(x)$, we then perform \textbf{token pruning}, discarding tokens with lower scores, thus completing the first carving stage. Assuming the overall goal is to reduce the number of visual tokens from $L_v$ to $L_{vc}$, with the merging proportion in the second stage denoted as $\rho$, this first stage reduces the number of tokens from $L_v$ to $L_{vc} \times (1+\rho)$.

\begin{table*}[t]
\centering
\renewcommand{\arraystretch}{1.1}
\resizebox{\textwidth}{!}{%
\fontsize{8pt}{10}\selectfont
\begin{tabularx}{\textwidth}{l| *{8}{>{\centering\arraybackslash}X}}
\hline
Method & MME\cite{tokencarve_1_mme} & MMB\cite{tokencarve_2_mmb} & POPE\cite{tokencarve_3_pope} & SQA$^{\text{IMG}}$\cite{tokencarve_4_sqa} & VizWiz\cite{tokencarve_5_vizwiz} & VQA$^{\text{Text}}$\cite{tokencarve_6_textvqa} & VQA$^{\text{V2}}$\cite{tokencarve_7_vqav2} & MMVet\cite{tokencarve_8_mmvet} \\
\hline
\cellcolor{gray!20}LLaVA-1.5-7B & \multicolumn{8}{c}{\cellcolor{gray!20}Upper Bound, 576 Tokens (\textbf{100\%)}} \\
\textbf{Vanilla} & 1868 & 64.6 & 86.1 & 69.5 & 50 & 58.2 & 78.5 & 31 \\
\hline  
\cellcolor{gray!20} & \multicolumn{8}{c}{\cellcolor{gray!20}Retain 192 Tokens \textcolor{black}(\textcolor{self_green}{{$\downarrow$} 66.7\%})} \\
\textbf{FastV}~\cite{tokencarve_9_fastv}  & \cellcolor{level_1}1865 & \cellcolor{level_1}64.3 & 78.2 & 68.9 & \cellcolor{level_2}50.9 & \cellcolor{level_2}58.0 & 75.9 & 28.9 \\
\textbf{SparseVLM}~\cite{tokencarve_11_sparsevlm} & 1721 & 62.5 & 83.6 & \cellcolor{level_2}69.1 & 50.5 & 56.1 & 75.6 & \cellcolor{level_2}31.5 \\
\textbf{MustDrop}~\cite{tokencarve_12_mustdrop} & 1787 & 62.3 & - & \cellcolor{level_1}69.2 & \cellcolor{level_1}51.4 & 56.5 & 76.0 & - \\
\textbf{VisionZip}~\cite{tokencarve_13_visionzip} & - & 63.0 & \cellcolor{level_1}85.3 & 68.9 & - & 57.3 & \cellcolor{level_2}76.8 & \cellcolor{level_1}31.7 \\
\textbf{TokenCarve (Ours)} & \cellcolor{level_2}1830 & 63.0 & \cellcolor{level_2}84.9 & \cellcolor{level_2}69.1 & \cellcolor{level_2}50.9 & \cellcolor{level_1}58.4 & \cellcolor{level_1}78.0 & 30.4 \\
\hline
\cellcolor{gray!20} & \multicolumn{8}{c}{\cellcolor{gray!20}Retain 128 Tokens \textcolor{black}(\textcolor{self_green}{{$\downarrow$} 77.8\%})} \\
\textbf{FastV}~\cite{tokencarve_9_fastv} & \cellcolor{level_2}1782 & \cellcolor{level_1}63.7 & 72.1 & \cellcolor{level_1}69.0 & 51.3 & \cellcolor{level_2}57.0 & 73.2 & 26.7 \\
\textbf{SparseVLM}~\cite{tokencarve_11_sparsevlm} & 1696 & 60.0 & 80.5 & 67.1 & \cellcolor{level_2}51.4 & 54.9 & 73.8 & \cellcolor{level_2}30.0 \\
\textbf{MustDrop}~\cite{tokencarve_12_mustdrop} & 1745 & 61.1 & - & 68.5 & \cellcolor{level_1}52.1 & 56.3 & 74.6 & - \\
\textbf{VisionZip}~\cite{tokencarve_13_visionzip} & - & 62.0 & \cellcolor{level_2}83.2 & \cellcolor{level_2}68.9 & - & 56.8 & \cellcolor{level_2}75.6 & \cellcolor{level_1}32.6 \\
\textbf{TokenCarve (Ours)} & \cellcolor{level_1}1829 & \cellcolor{level_2}62.7 & \cellcolor{level_1}84.5 & \cellcolor{level_2}68.9 & 51.0 & \cellcolor{level_1}58.1 & \cellcolor{level_1}77.3 & 29.5 \\
\hline
\cellcolor{gray!20} & \multicolumn{8}{c}{\cellcolor{gray!20}Retain 64 Tokens \textcolor{black}(\textcolor{self_green}{{$\downarrow$} 88.9\%})} \\
\textbf{FastV}~\cite{tokencarve_9_fastv} & 1564 & 61.0 & 59.2 & \cellcolor{level_1}69.9 & \cellcolor{level_1}51.8 & 55.1 & 66.3 & 26.1 \\
\textbf{SparseVLM}~\cite{tokencarve_11_sparsevlm} & 1505 & 56.2 & 75.1 & 62.2 & 50.1 & 51.8 & 68.2 & 23.3 \\
\textbf{MustDrop}~\cite{tokencarve_12_mustdrop} & \cellcolor{level_2}1641 & 60.0 & - & 63.4 & 51.2 & 54.2 & 69.3 & - \\
\textbf{VisionZip}~\cite{tokencarve_13_visionzip} & - & \cellcolor{level_2}61.5 & \cellcolor{level_1}80.9 & 68.8 & - & \cellcolor{level_2}56.0 & \cellcolor{level_2}72.4 & \cellcolor{level_1}30.2 \\
\textbf{TokenCarve (Ours)} & \cellcolor{level_1}1754 & \cellcolor{level_1}62.0 & \cellcolor{level_2}79.9 & \cellcolor{level_2}69.7 & \cellcolor{level_2}51.4 & \cellcolor{level_1}57.0 & \cellcolor{level_1}74.8 & \cellcolor{level_2}29.3 \\
\hline
\end{tabularx}
}
\caption{Evaluation of TokenCarve on the LLaVA-1.5-7B model across eight datasets under three visual token compression levels (192, 128, and 64). The vanilla configuration uses 576 tokens. We compare TokenCarve against both published methods (e.g., FastV~\cite{tokencarve_9_fastv}, VisionZip~\cite{tokencarve_13_visionzip}) and preprints (e.g., MustDrop~\cite{tokencarve_12_mustdrop}, SparseVLM~\cite{tokencarve_11_sparsevlm}). \textcolor[HTML]{EDC1C0}{Red} denotes the best performance, and \textcolor[HTML]{F6DFC1}{orange} denotes the second best. TokenCarve generally achieves top or near-top results across most benchmarks, with its advantages becoming more pronounced under extreme compression. Notably, it performs especially well on OCR-heavy tasks such as VQA$^{\text{Text}}$ and VQA$^{\text{V2}}$.}

\label{tab:performance7b}
\vspace{-4mm}
\end{table*}

\subsubsection{Carving Stage II: Using an Information-Preserving Guided Selection Strategy for Token Merge}
In the second stage of token carving, we have already pruned out a portion of tokens using the IPGS strategy, leaving behind those that are relatively more important. Therefore, a more fine-grained token carving method is needed at this stage. To further preserve information, we adopt a token merging~\cite{tokencarve_10_tome} approach in Carving Stage II that aligns with our information-preserving principle. Unlike conventional merging strategies that group tokens solely based on their positions, our approach utilizes the token scores provided by IPGS to divide the tokens into two groups according to their importance. Specifically, we partition the tokens as follows:
\begingroup
\setlength{\abovedisplayskip}{2.5pt}    
\setlength{\belowdisplayskip}{2pt}    
\begin{equation}
\hspace*{-0.5em} 
\begin{aligned}
\text{SetA} &= \text{IndexSort}(Z_{\text{visual}}^2)\Big[ : \frac{L_{vc}(1+\rho)}{2} \Big], \\
\text{SetB} &= \text{IndexSort}(Z_{\text{visual}}^2)\Big[ \frac{L_{vc}(1+\rho)}{2} : L_{vc}(1+\rho) \Big],
\end{aligned}
\label{eq:3-9}
\end{equation}
\endgroup
where \textit{IndexSort} returns the indices of the sequence sorted in descending order.

After obtaining SetA and SetB, we normalize the corresponding tokens using Equation (2) and perform matrix multiplication to compute the similarity between tokens in SetA and SetB as follows:
\begin{equation}
\text{Simi}_{AB} = \text{Norm}(Z^2)[\text{SetB}] \times \text{Norm}(Z^2)[\text{SetA}],
\label{eq:3-10}
\end{equation}
where $\times$ denotes matrix multiplication.

Finally, the top \(L_{vc} \times \rho\) tokens in SetB—those exhibiting the highest similarity scores with tokens in SetA—are merged into SetA using an average weighting scheme (as employed by ToMe~\cite{tokencarve_10_tome}), while all remaining tokens remain intact. This process completes the second carving stage, reducing the number of visual tokens from \(L_v\) to \(L_{vc}\). Section~\ref{Effectiveness Analysis} further validates the effectiveness of TokenCarve in visual token compression and investigates the underlying factors contributing to its performance.

\section{Experiments}
\subsection{Evaluation Tasks}

To assess the effectiveness of our method on image understanding tasks, we conduct experiments on eight widely used benchmarks including MME~\cite{tokencarve_1_mme}, MMBench (MMB)~\cite{tokencarve_2_mmb}, POPE~\cite{tokencarve_3_pope}, SQA$^{\text{IMG}}$~\cite{tokencarve_4_sqa}, VizWiz~\cite{tokencarve_5_vizwiz}, VQA$^{\text{Text}}$~\cite{tokencarve_6_textvqa}, VQA$^{\text{V2}}$~\cite{tokencarve_7_vqav2} and MMVet~\cite{tokencarve_8_mmvet}. At the same time,  we compare our method with the existing sota methods such as FastV~\cite{tokencarve_9_fastv}, SparseVLM~\cite{tokencarve_11_sparsevlm}, MustDrop~\cite{tokencarve_12_mustdrop} and VisionZip~\cite{tokencarve_13_visionzip}, which mainly utilize attention scores to compress redundant visual tokens in MLLMs. 

\subsection{Implementation Details}
To further validate the generalizability of our method, we test TokenCarve on various open-source MLLMs with different architectures. We conduct experiments mainly on LLaVA-1.5-7B frameworks. Experiments on LLaVA-1.5-13B further verify the effectiveness of our method on larger scale model. All experiments are done on a single NVIDIA Tesla V100 with 32GB.

\subsection{Main Results}
\textbf{Results on LLaVA-1.5-7B.} In~\cref{tab:performance7b}, we verify our method on LLaVA-1.5-7B with eight datasets, respectively. The comparisons included formally published methods(~\cite{tokencarve_9_fastv,tokencarve_13_visionzip}) as well as arxiv preprints(~\cite{tokencarve_12_mustdrop,tokencarve_11_sparsevlm}). Following the setup in~\cite{tokencarve_9_fastv}, we use three visual token count configurations (192, 128, and 64) to evaluate the advantages of our proposed visual q\&a. Overall, our method achieves optimal or suboptimal results on almost all datasets, demonstrating the powerful competitiveness of our method. \textbf{As the number of visual tokens decreases, the performance advantages of our approach become increasingly apparent, which fully verifies that TokenCarve effectively mitigates performance degradation during token compression}. The visual q\&a task datasets (such as VQA$^{\text{Text}}$ and VQA$^{\text{V2}}$) contain a large number of OCR tasks that are sensitive to the loss of informativeness due to token compression. Our method performs excellently on these visual q\&a tasks and achieves optimality in three visual token count configurations compared to all existing methods. An interesting phenomenon is that the SQA and VizWizVQA datasets are not sensitive to token compression rate, which is present in all visual token compression methods. About results across more visual token count settings and additional datasets, please refer to the supplementary materials.

\begin{table}[t]
\centering

\renewcommand{\arraystretch}{1.1}
\resizebox{\linewidth}{!}{%
\fontsize{8pt}{10}\selectfont
\begin{tabularx}{\linewidth}{l *{4}{>{\centering\arraybackslash}X}}
\hline
\multicolumn{1}{l|}{\textbf{Method}} & MME & POPE & VQA$^{\text{Text}}$ & MMVet \\
\hline
\multicolumn{1}{l|}{\cellcolor{gray!20}LLaVA-1.5-13B} & \multicolumn{4}{c}{\cellcolor{gray!20}Upper Bound, 576 Tokens (\textbf{100\%)}} \\
\multicolumn{1}{l|}{Vanilla} & 1531/296 & 86.4 & 61.2 & 36.1 \\
\hline   
\multicolumn{1}{l|}{\cellcolor{gray!20}} & \multicolumn{4}{c}{\cellcolor{gray!20}Retain 192 Tokens \textcolor{black}(\textcolor{self_green}{{$\downarrow$} 66.7\%})} \\
\multicolumn{1}{l|}{FastV}  & 1493/296 & 83.3 & 60.5 & \textbf{35} \\
\multicolumn{1}{l|}{TokenCarve (Ours)} & \textbf{1535/311} & \textbf{86.8} & \textbf{60.8} & 34.4 \\
\hline
\multicolumn{1}{l|}{\cellcolor{gray!20}} & \multicolumn{4}{c}{\cellcolor{gray!20}Retain 128 Tokens \textcolor{black}(\textcolor{self_green}{{$\downarrow$} 77.8\%})} \\
\multicolumn{1}{l|}{FastV} & 1484/306 & 78 & 59.7 & 34 \\
\multicolumn{1}{l|}{TokenCarve (Ours)} & \textbf{1507/315} & \textbf{85.8} & \textbf{60.6} & \textbf{35.6} \\
\hline
\multicolumn{1}{l|}{\cellcolor{gray!20}} & \multicolumn{4}{c}{\cellcolor{gray!20}Retain 64 Tokens \textcolor{black}(\textcolor{self_green}{{$\downarrow$} 88.9\%})} \\
\multicolumn{1}{l|}{FastV} & 1373/284 & 69.2 & 56.1 & 31.2\\
\multicolumn{1}{l|}{TokenCarve (Ours)} & \textbf{1483/285} & \textbf{83.7} & \textbf{59.3} & \textbf{33.9} \\
\hline
\end{tabularx}
}
\caption{Performance comparison of various token compression methods applied to the LLaVA-1.5-13B model. On three datasets, TokenCarve exhibits outstanding performance, with its advantages becoming increasingly pronounced under more extreme token reduction conditions.}
\label{tab:performance13b}
\vspace{-4mm}
\end{table}

\textbf{Results on LLaVA-1.5-13B.} To further demonstrate the effectiveness of our proposed TokenCarve, we apply it to the larger model LLaVA-1.5-13B, as shown in~\cref{tab:performance13b}. It can be found that our method overwhelmingly outperforms FastV on the four datasets. In MME dataset, TokenCarve scored significantly higher than fastV on both Perception / Cognition parts. 
Obviously, in conditions where fewer visual tokens are retained, the more significant is the advantage of our approach compared to FastV. In POPE dataset, when 64 visual tokens are retained, FastV's performance drops by 17.2, yet TokenCarve drops by only 2.7. These results validate that \textbf{TokenCarve maintains strong performance even under extremely low token retention rates}, demonstrating its effectiveness across different model sizes and confirming its robustness across architectures.

\begin{table}[t]
\centering
\resizebox{\columnwidth}{!}{%
\fontsize{40pt}{80}\selectfont
\begin{tabular}{@{}ccccccccc@{}}
\hline
\multicolumn{1}{l|}{{\textbf{Method/Visual token}}} & \textbf{333}     & \textbf{288}     & \textbf{192}     & \textbf{144}     & \textbf{128}     & \textbf{72}      & \textbf{64}      & \textbf{Avg Drop}                  \\ \hline
\rowcolor[HTML]{E6E6E6} 
\multicolumn{1}{c|}{\textbf{LLaVA-1.5-7B} }           & \multicolumn{8}{c}{\cellcolor[HTML]{E6E6E6}\textbf{MME (Baseline 1868)}}                                                                                                 \\
\multicolumn{1}{c|}{}                             & 1847             & \textbf{1872}    & \textbf{1840}    & 1813             & 1793             & 1681             & 1684             &                                    \\
\multicolumn{1}{c|}{\multirow{-2}{*}{\fontsize{55pt}{80}\selectfont AS} }    & -1.12\%          & \textbf{+0.21\%} & \textbf{-1.5\%}  & -2.94\%          & -4.01\%          & -10.01\%         & -9.85\%          & \multirow{-2}{*}{-4.18\%}          \\ \hline
\multicolumn{1}{c|}{}                             & 1829             & 1817             & 1810             & 1810             & 1796             & 1729             & 1705             &                                    \\
\multicolumn{1}{c|}{\multirow{-2}{*}{\fontsize{55pt}{80}\selectfont ICS}}     & -2.09\%          & -2.73\%          & -3.1\%           & -3.1\%           & -3.85\%          & -7.44\%          & -8.73\%          & \multirow{-2}{*}{-4.44\%}          \\ \hline
\multicolumn{1}{c|}{}                             & \textbf{1867}    & 1867             & 1832             & \textbf{1821}    & \textbf{1827}    & \textbf{1767}    & \textbf{1754}    &                                    \\
\multicolumn{1}{c|}{\multirow{-2}{*}{\fontsize{55pt}{80}\selectfont TokenCarve}} & \textbf{-0.05\%} & -0.05\%          & -1.93\%          & \textbf{-2.52\%} & \textbf{-2.19\%} & \textbf{-5.41\%} & \textbf{-6.1\%}  & \multirow{-2}{*}{\textbf{-2.61\%}} \\ \hline
\rowcolor[HTML]{E6E6E6} 
\multicolumn{1}{c|}{\textbf{LLaVA-1.5-7B}}            & \multicolumn{8}{c}{\cellcolor[HTML]{E6E6E6}\textbf{POPE (Baseline 86.1)}}                                                                                                \\
\multicolumn{1}{c|}{}                             & 86.9             & 86.6             & 84.7             & 82.4             & 81.5             & 76.1             & 74.5             &                                    \\
\multicolumn{1}{c|}{\multirow{-2}{*}{\fontsize{55pt}{80}\selectfont AS}}     & +0.93\%          & +0.58\%          & -1.63\%          & -4.3\%           & -5.34\%          & -11.61\%         & -13.47\%         & \multirow{-2}{*}{-4.98\%}          \\ \hline
\multicolumn{1}{c|}{}                             & \textbf{87.1}    & \textbf{87.1}    & \textbf{87.3}    & \textbf{87.1}    & \textbf{86.8}    & \textbf{84.8}    & \textbf{83.9}    &                                    \\
\multicolumn{1}{c|}{\multirow{-2}{*}{\fontsize{55pt}{80}\selectfont ICS}}     & \textbf{+1.16\%} & \textbf{+1.16\%} & \textbf{+1.39\%} & \textbf{+1.16\%} & \textbf{+0.81\%} & \textbf{-1.51\%} & \textbf{-2.56\%} & \multirow{-2}{*}{\textbf{+0.23\%}} \\ \hline
\multicolumn{1}{c|}{}                             & 87               & 86.7             & 86.4             & 86.2             & 86.1             & 83.7             & 82.9             &                                    \\
\multicolumn{1}{c|}{\multirow{-2}{*}{\fontsize{55pt}{80}\selectfont TokenCarve}} & +1.05\%          & +0.7\%           & +0.35\%          & +0.12\%          & 0.00\%           & -2.79\%          & -3.72\%          & \multirow{-2}{*}{-0.61\%}          \\ \hline
\rowcolor[HTML]{E6E6E6} 
\multicolumn{1}{c|}{\textbf{LLaVA-1.5-7B}}            & \multicolumn{8}{c}{\cellcolor[HTML]{E6E6E6}\textbf{TextVQA (Baseline 58.2)}}                                                                                             \\
\multicolumn{1}{c|}{}                             & 58               & 58               & 57.9             & 57.3             & 57.2             & 56               & 55.7             &                                    \\
\multicolumn{1}{c|}{\multirow{-2}{*}{\fontsize{55pt}{80}\selectfont AS}}     & -0.34\%          & -0.34\%          & -0.51\%          & -1.52\%          & -1.69\%          & -3.72\%          & -4.22\%          & \multirow{-2}{*}{-1.76\%}          \\ \hline
\multicolumn{1}{c|}{}                             & \textbf{58.4}    & \textbf{58.6}    & 58               & 57.7             & 57.2             & 54.6             & 54.1             &                                    \\
\multicolumn{1}{c|}{\multirow{-2}{*}{\fontsize{55pt}{80}\selectfont ICS}}     & \textbf{+0.34\%} & \textbf{+0.68\%} & -0.34\%          & -0.84\%          & -1.69\%          & -6.08\%          & -6.93\%          & \multirow{-2}{*}{-2.12\%}          \\ \hline
\multicolumn{1}{c|}{}                             & 58.2             & 58.1             & \textbf{58.4}    & \textbf{58.2}    & \textbf{58.1}    & \textbf{57.2}    & \textbf{57.1}    &                                    \\
\multicolumn{1}{c|}{\multirow{-2}{*}{\fontsize{55pt}{80}\selectfont TokenCarve}} & 0.00\%           & -0.17\%          & \textbf{+0.34\%} & \textbf{0.00\%}  & \textbf{-0.17\%} & \textbf{-1.69\%} & \textbf{-1.86\%} & \multirow{-2}{*}{\textbf{-0.51\%}} \\ \hline
\end{tabular}
}
\caption{Comparison of token selection strategies for visual token compression on LLaVA-1.5-7B across three datasets. Results are shown for various token counts using Attention Score (AS) and Information Contribution Score (ICS). While ICS performs exceptionally on POPE, it underperforms on others. In contrast, TokenCarve achieves the lowest average degradation and more stable results, underscoring the critical role of effective token selection.}

\label{tab:ablation1}
\vspace{-6mm}
\end{table}

\subsection{Ablation Study}
\textbf{Effect of ICS.} To understand the effectiveness of minimizing Information Loss, we evaluate three datasets (i.e. MME, POPE and TextVQA) in~\cref{tab:ablation1}, where AS means attention score and ICS means information contribution score. The performance of ICS on the POPE dataset is extremely bright and even achieves better results compared to the joint approach, reflecting our significance of minimizing information loss. But its performance on other datasets is the worst. Overall, the joint selection strategy (TokenCarve) is the most stable and performs better in general, especially when the number of tokens is fewer. Therefore the joint selection strategy is the best choice in most cases.

\begin{figure}[h!]
\vspace{-2mm}
  \centering
  \includegraphics[width=\linewidth]{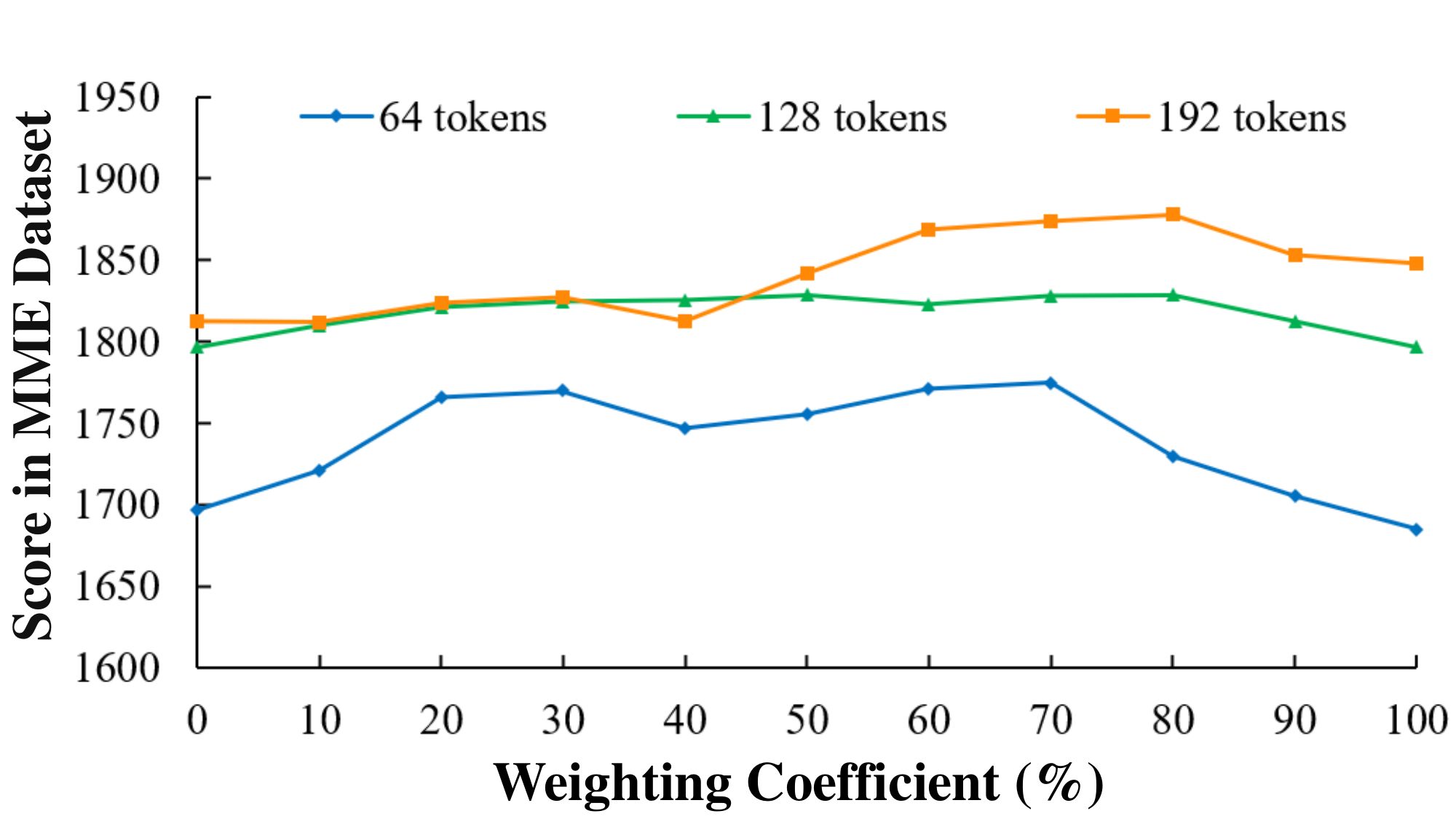}
\caption{Impact of weighting coefficient ($\lambda$) on model performance. The results show that extreme values of $\lambda$ at both ends lead to performance degradation, highlighting the importance of the combined score.}
  \label{fig:ablation2}  
  \vspace{-2mm}
\end{figure}

\textbf{Necessity of Weighting Coefficient.} To evaluate the impact of our selection strategy in visual token compression, we analyze the effect of weighting coefficient variation on MLLMs' performance in~\cref{fig:ablation2}. The results reveal a consistent trend across different visual token count configurations: when the weighting coefficient approaches either extreme (0\% or 100\%), model performance declines. This observation highlights the crucial role of the combined score. TokenCarve, by leveraging the combined score, effectively prunes redundant visual tokens while preserving essential visual information, ensuring a balanced and robust compression strategy.

\begin{figure}[h]
\vspace{-2mm}
  \centering
  \includegraphics[width=\linewidth]{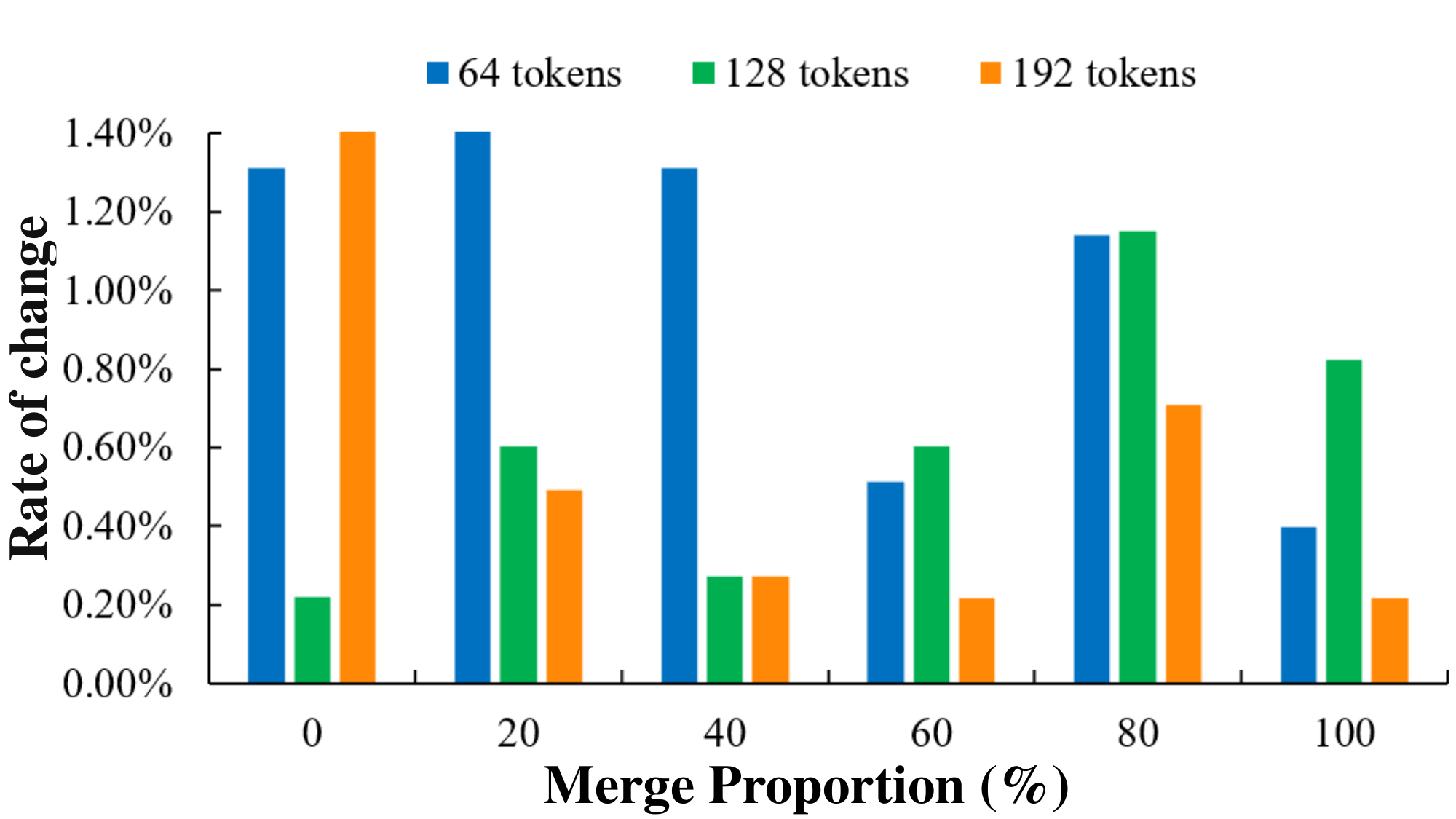}
\caption{Impact of merge proportion ($\rho$) on model performance. The results indicate that TokenCarve remains robust across different values of $\rho$, with performance variations staying within 1.5\%.}

  \label{fig:ablation1} 
  \vspace{-4mm}
\end{figure}

\textbf{Sensitivity of Merge Proportion.} In~\cref{fig:ablation1}, we examine the impact of different merge proportions on performance, using LLaVA-1.5-7B as the base model and MME as the dataset. We compute the rate of change in MLLM performance across various visual token retention levels relative to the 50\% retention in the main experiments. As the merge proportion increases, model performance remains within a 1.5\% variation across all three visual token count configurations. This clearly demonstrates that TokenCarve is insensitive to merge proportion changes, confirming the robustness of our method.

\begin{figure}[h]
  \centering
  \includegraphics[width=\linewidth]{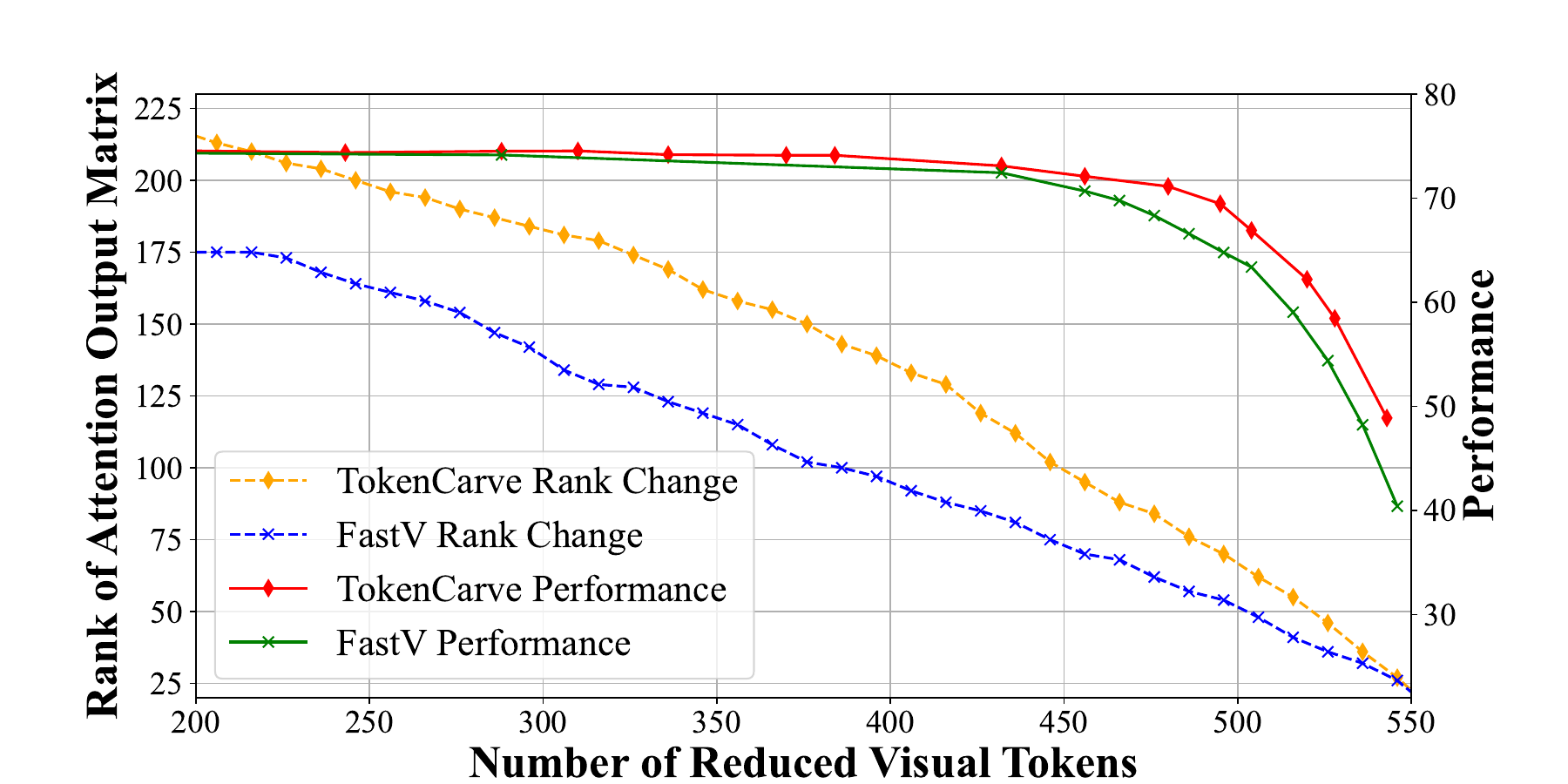}
\caption{This figure illustrates that TokenCarve better preserves the rank of the Attention Output Matrix during visual token compression, which underlies its effectiveness. Specifically, the dashed lines represent how the matrix rank changes as the number of visual tokens decreases for TokenCarve and FastV, respectively, while the solid lines correspond to the actual performance variations of TokenCarve and FastV with reduced token counts.}

  \label{fig:Figure_6}  
  \vspace{-4mm}
\end{figure}

\subsection{Effectiveness Analysis}
\label{Effectiveness Analysis}
As shown in Figure~\ref{fig:Figure_6}, we conducted a detailed analysis of the effectiveness of TokenCarve in visual token compression. As clearly illustrated by the two dashed curves in the lower-left region (the orange dashed curve representing TokenCarve and the blue dashed curve representing FastV), as the number of visual tokens decreases, the orange dashed curve corresponding to TokenCarve consistently remains higher than the blue dashed curve of FastV. This indicates that TokenCarve better preserves the rank of the Attention Output Matrix, effectively retaining more information from the original visual token sequences. Additionally, as demonstrated by the performance curves on the right axis, TokenCarve consistently maintains superior performance compared to FastV, especially under conditions of significant token reduction. This phenomenon validates our key insight from Figure~\ref{fig:3-1} and further confirms that \textbf{TokenCarve efficiently preserves the rank of the Attention Output Matrix, thereby maintaining the information quantity of the token sequence, which fundamentally underpins its effectiveness.}

\subsection{CUDA Time Test}
In~\cref{tab:speed}, we assess the practical acceleration of TokenCarve on a single NVIDIA Tesla V100 (32GB) using \texttt{torch.profiler}, evaluating 100 sampled instances from the TextVQA dataset with execution times averaged across samples. As TokenCarve compresses visual tokens, it significantly reduces both inference latency and KV cache occupancy. Specifically, decreasing the visual token count from 576 to 128 reduces inference latency from 1.124 s to 0.911 s, achieving a 1.23× speedup, while KV cache usage drops from 100\% to 36\%. Additionally, the model maintains high performance stability, with accuracy decreasing only slightly from 58.2 to 58.1 across different token configurations. Further compressing the visual token count to 64 results in a performance drop of 1.2, while KV cache usage is reduced to 26.9\%, achieving an even greater acceleration of 1.33×. These results confirm the practical effectiveness of TokenCarve, demonstrating its capability to reduce computational and storage demands, thereby enhancing the real-world deployability of MLLMs.

\begin{table}[h]
\centering
\resizebox{\columnwidth}{!}{%
\fontsize{8pt}{10}\selectfont
\begin{tabular}{ccccc}
\hline
\multicolumn{1}{c|}{\textbf{Visual token}} & \textbf{576} & \textbf{192} & \textbf{128} & \textbf{64} \\ \hline
\multicolumn{1}{c|}{\textbf{Performance}}    & 58.2       & 58.4       & 58.1       & 57.0       \\ 
\multicolumn{1}{c|}{\textbf{KV Cache (\%)}}    & 100\%      & 45.1\%     & 36\%       & 26.9\%     \\
\multicolumn{1}{c|}{\textbf{Latency (s)}}      & \makecell[c]{1.124 \\ (1.00$\times$)}     & \makecell[c]{0.974 \\ (1.15$\times$)}      & \makecell[c]{0.911 \\ (1.23$\times$)}      & \makecell[c]{0.845 \\ (1.33$\times$)}      \\ \hline
\end{tabular}
}
\caption{Performance, latency, and KV cache usage comparison under different visual token configurations.}
\label{tab:speed}
\vspace{-4mm}
\end{table}

\section{Conclusion}

In this study, we investigate the relationship between MLLM performance and information retention in the attention output matrix, revealing that performance degradation closely follows the loss of information during token compression. Motivated by this insight, we propose TokenCarve—a training-free, plug-and-play, two-stage token compression framework that maintains robust performance even at extremely low token compression ratios. By employing an Information-Preservation-Guided Selection (IPGS) strategy to prune and merge tokens, TokenCarve reduces the visual token count to 22.2\% of the original while achieving a 1.23× inference speedup, a 64\% reduction in KV cache storage, and only a 1.54\% drop in accuracy.

Despite these promising results, the actual acceleration remains constrained by engineering limitations. For example, retaining the original positional relationships of sparsified tokens necessitates repeated computation of rotary positional encodings using the original position IDs, which restricts the model’s effective speedup. Our future work focuses on optimizing these practical aspects to achieve more significant inference speed improvements in real-world applications.
{
    \small
    \bibliographystyle{ieeenat_fullname}
    \bibliography{main}
}

\clearpage
\appendix
\maketitlesupplementary
\vspace{20pt}

\section{Visualization Result Analysis}

\begin{figure*}[h]
\vspace{-3mm}
  \centering
  \includegraphics[width=0.95\linewidth]{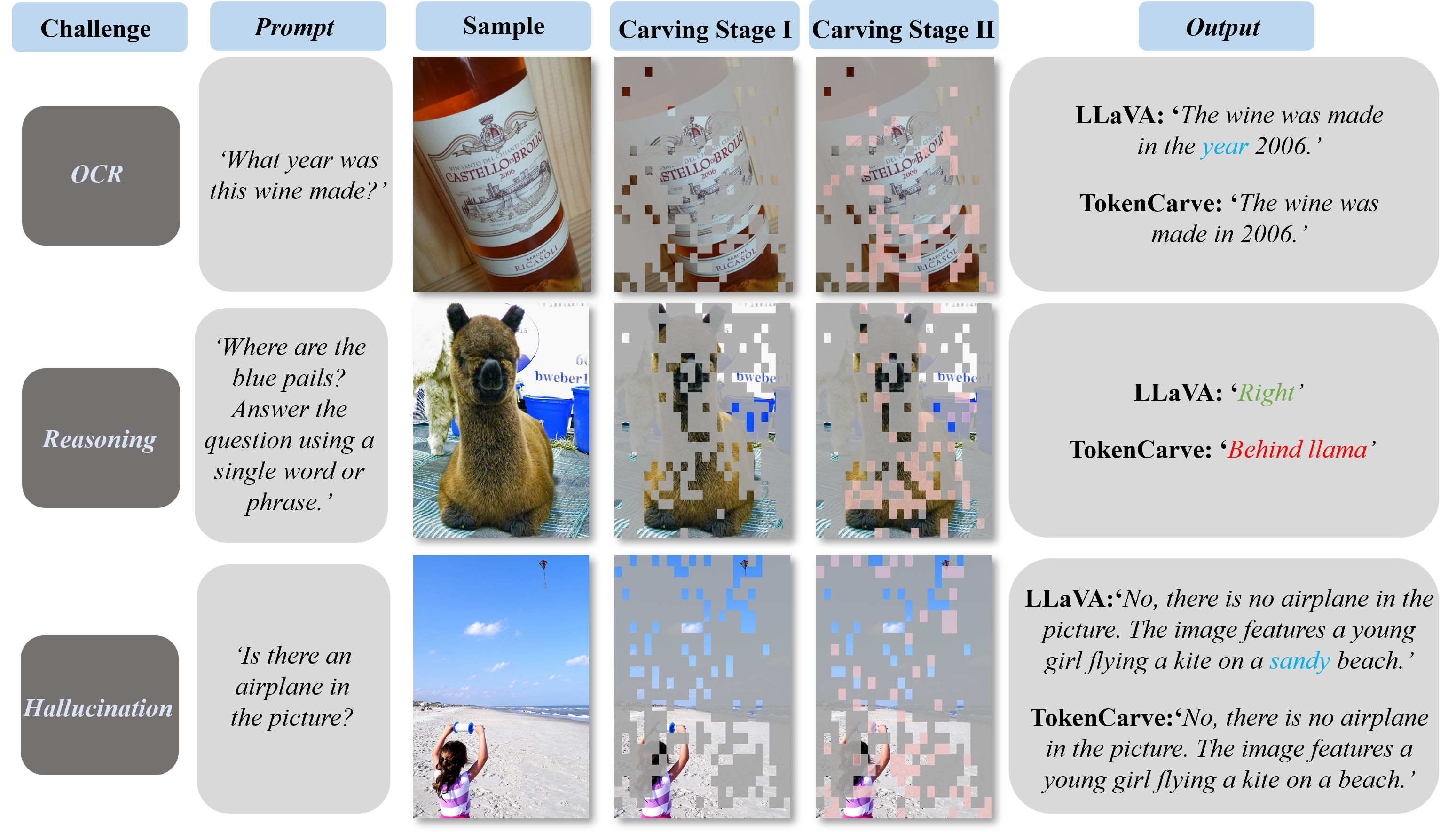}
\caption{Visualization comparison between TokenCarve and original LLaVA across OCR, reasoning, and hallucination tasks.}
\label{fig:visualization}
\vspace{-3mm}
\end{figure*}

In this section, we thoroughly analyze the visualization results of the TokenCarve method, which reduces the number of visual tokens from 576 down to 128. We conduct visualization experiments across three key tasks: Optical Character Recognition (OCR), reasoning, and hallucination. Comparisons are made with the original LLaVA model, where tokens remain uncompressed. In the OCR task, TokenCarve continues to distinctly focus on regions containing the necessary textual information after token compression. As demonstrated in~\cref{fig:visualization}, the extracted text from TokenCarve is highly similar to that obtained by the original LLaVA model, and both methods accurately extract the target text. For the reasoning task, TokenCarve maintains attention on critical details mentioned in the prompt even after compressing tokens. Interestingly, the final answers from the TokenCarve model sometimes outperform those of the original uncompressed LLaVA model. In the hallucination task, TokenCarve consistently retains its attention on the key spatial details indicated by the prompt, avoiding hallucinations caused by token loss. Furthermore, the final responses from TokenCarve remain comparable to those generated by the original LLaVA model without token compression.

\section{Attention Output Matrix Rank}
\begin{figure*}[ht]
  \centering
  \includegraphics[width=0.9\linewidth]{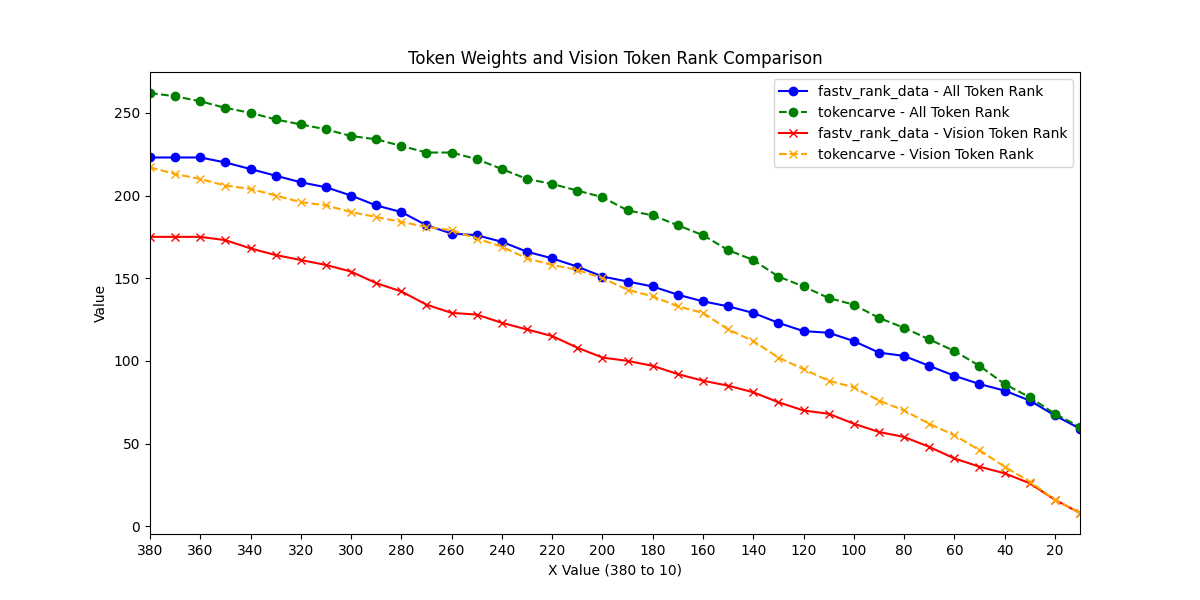}
  \vspace{-5mm}
\caption{Attention output matrix rank.}
    
  \label{fig:rank}  
  \vspace{-6mm}
\end{figure*}

We analyze the effect of the decrease in the number of Token in LLaVA on the rank of the Token matrix in~\cref{fig:rank}. It can be found that as the number of all Token or the number of visual tokens decreases, their rank decreases accordingly, which is consistent with the decrease in their performance. However, TokenCarve-guided visual token compression always has a higher matrix rank than FastV for the same compression rate, which further proves that our method is able to retain more visual information.

\section{Datasets}
\begin{enumerate}
    \item MME (Multimodal Model Evaluation)\cite{tokencarve_1_mme} dataset is a systematic framework designed to evaluate the comprehensive performance of multimodal large language models across two dimensions: perceptual capabilities (e.g., visual information recognition) and cognitive abilities (e.g., reasoning, computation, translation). The detailed results on this dataset are in~\cref{tab:Result on MME}.
    \item MMB (Multimodal Model Benchmark)\cite{tokencarve_2_mmb} dataset serves as a benchmark for assessing the holistic capabilities of multimodal foundation models in diverse scenarios. It focuses on three tiers: low-level perception (e.g., color and spatial recognition), fine-grained perception (e.g., text recognition and object relationship inference), and higher-order reasoning (e.g., logical reasoning and multimodal dialogue). The detailed results on this dataset are in~\cref{tab:Result on MMB}.
    \item MMB-CN\cite{tokencarve_2_mmb}, the Chinese adaptation of the MMB dataset, specializes in evaluating multimodal tasks in Chinese linguistic contexts. It primarily measures multimodal foundation models' perceptual and reasoning capabilities in Chinese scenarios, facilitating cross-linguistic performance analysis for researchers. The detailed results on this dataset are in~\cref{tab:Result on MMB-CN}.
    \item POPE (Polling-based Object Probing Evaluation)\cite{tokencarve_3_pope} is a diagnostic dataset for quantifying object hallucination in large vision-language models. It employs a polling mechanism to detect instances where models erroneously "perceive" non-existent objects in images. The detailed results on this dataset are in~\cref{tab:Result on POPE} (LLaVA-7B) and~\cref{tab:Result on POPE 13B} (LLaVA-13B).
    \item SQA (Science Question Answering)\cite{tokencarve_4_sqa} is a multimodal scientific QA dataset comprising 21,208 multiple-choice questions derived from K-12 science curricula. Spanning chemistry, biology, physics, and other disciplines, each question integrates multimodal contextual information, correct answers, background knowledge, and detailed explanations. The detailed results on this dataset are in~\cref{tab:Result on SQAIMG}.
    \item VizWiz\cite{tokencarve_5_vizwiz} serves as an evaluation benchmark for multimodal foundation models in visual question answering tasks. It specifically measures three critical competencies: real-world image comprehension, complex problem-solving capacity, and identification of unanswerable questions. The detailed results on this dataset are in~\cref{tab:Result on VizWizVQA}.
    \item TextVQA\cite{tokencarve_6_textvqa} focuses on evaluating visual question answering models' optical character recognition (OCR) capabilities and text comprehension in visual contexts. This dataset requires multimodal reasoning to address questions directly tied to textual elements within images. The detailed results on this dataset are in~\cref{tab:Result on TextVQA} (LLaVA-7B) and~\cref{tab:Result on TextVQA 13B} (LLaVA-13B).
    \item VQAV2\cite{tokencarve_7_vqav2} is a benchmark for assessing multimodal foundation models' proficiency in visual understanding, linguistic processing, and commonsense reasoning. Its primary focus lies in evaluating models' ability to accurately interpret visual content and generate contextually appropriate responses. The detailed results on this dataset are in~\cref{tab:Result on VQAV2}.
    \item MM-Vet (Multimodal Versatile Evaluation Toolkit)\cite{tokencarve_8_mmvet} is a comprehensive benchmark designed to evaluate multimodal foundation models through complex cross-modal tasks. It systematically measures seven core visual-language capabilities: recognition, OCR, knowledge integration, language generation, spatial reasoning, mathematical computation, and sequential image-text understanding. The detailed results on this dataset are in~\cref{tab:Result on MM-Vet} (LLaVA-7B) and~\cref{tab:Result on MM-Vet 13B} (LLaVA-13B).
    \item OCR-VQA\cite{tokencarve_15_ocrvqa} evaluates multimodal foundation models' performance in text recognition (OCR), visual comprehension, and multimodal reasoning, with particular emphasis on accurate extraction and utilization of textual information embedded in images. The detailed results on this dataset are in~\cref{tab:Result on OCRVQA}.
    \item AOK-VQA (All-in-One Knowledge VQA)\cite{tokencarve_40_okvqa} is a diagnostic dataset for visual question answering that emphasizes models' ability to integrate visual content with external world knowledge for complex reasoning. Its questions necessitate extensive commonsense reasoning and world knowledge beyond simple knowledge base retrieval. The detailed results on this dataset are in~\cref{tab:Result on AOKVQA}.
\end{enumerate}

\section{More results on different datasets}
\subsection{Results on LLaVA-7B}

We give more detailed results on each dataset.TokenCarve shows better performance on different datasets with different visual token compression rates. We find that our method performs better compared to FastV when the visual token compression rate is high. On some datasets (e.g., VizWiz), the model's performance does not change much as the number of visual tokens decreases, so there is little difference between TokenCarve and fastv. More detailed results are given below.

\begin{table}[h!]
\resizebox{\columnwidth}{!}{%
\fontsize{7pt}{10}\selectfont
\begin{tabular}{c|cccccccc}
\hline
\textbf{Method/Visual token}        & \textbf{576}  & \textbf{333}  & \textbf{288}  & \textbf{192}  & \textbf{144}  & \textbf{128}  & \textbf{72}   & \textbf{64}   \\ \hline
\textbf{Fastv}      & \textbf{1868} & \textbf{1870} & \textbf{1869} & \textbf{1865} & 1800          & 1782          & 1612          & 1564          \\
\textbf{TokenCarve} & -             & 1855          & 1860          & 1830          & \textbf{1821} & \textbf{1829} & \textbf{1770} & \textbf{1754} \\ \hline
\end{tabular}}
\caption{Result on MME}
\label{tab:Result on MME}
\vspace{-3mm}
\end{table}

\begin{table}[h!]
\resizebox{\columnwidth}{!}{%
\fontsize{7pt}{10}\selectfont
\begin{tabular}{c|cccccccc}
\hline
\textbf{Method/Visual token}        & \textbf{576}  & \textbf{333}  & \textbf{288}  & \textbf{192}  & \textbf{144}  & \textbf{128}  & \textbf{72}   & \textbf{64} \\ \hline
\textbf{Fastv}      & \textbf{64.6} & \textbf{64.7} & \textbf{64.3} & \textbf{64.3} & \textbf{64.1} & \textbf{63.7} & 61.7          & 61          \\
\textbf{TokenCarve} & -             & 63.8          & 64            & 63            & 62.3          & 62.7          & \textbf{62.3} & \textbf{62} \\ \hline
\end{tabular}}
\caption{Result on MMB}
\label{tab:Result on MMB}
\vspace{-3mm}
\end{table}

\begin{table}[h!]
\resizebox{\columnwidth}{!}{%
\fontsize{7pt}{10}\selectfont
\begin{tabular}{c|cccccccc}
\hline
\textbf{Method/Visual token}     & \textbf{576} & \textbf{333} & \textbf{288} & \textbf{192} & \textbf{144} & \textbf{128} & \textbf{72} & \textbf{64} \\ \hline
\textbf{Fastv}      & \textbf{58.1}         & \textbf{58.5}         & \textbf{58.2}         & 57.5         & 57           & 56.4         & 54.1        & 52.8        \\
\textbf{TokenCarve} & -            & 57.6         & 57.9         & \textbf{58.4}         & \textbf{58.2}         & \textbf{57.9}         & \textbf{56}          & \textbf{55.8}        \\ \hline
\end{tabular}}
\caption{Result on MMB-CN}
\label{tab:Result on MMB-CN}
\vspace{-3mm}
\end{table}

\begin{table}[h!]
\resizebox{\columnwidth}{!}{%
\fontsize{7pt}{10}\selectfont
\begin{tabular}{c|cccccccc}
\hline
\textbf{Method/Visual token}                     & \textbf{576}  & \textbf{333}  & \textbf{288}  & \textbf{192}  & \textbf{144}  & \textbf{128}  & \textbf{72}   & \textbf{64}   \\ \hline
\textbf{Fastv (popular, F1)}      & \textbf{86.1} & 84.1          & 82.6          & 78.2          & 74.1          & 72.1          & 62.5          & 59.2          \\
\textbf{TokenCarve (popular, F1)} & -             & \textbf{85.8} & \textbf{85.4} & \textbf{84.9} & \textbf{84.6} & \textbf{84.5} & \textbf{81.2} & \textbf{79.9} \\
\textbf{Fastv (random, F1)}       & \textbf{87.3} & 84.7          & 83.3          & 78.6          & 74.3          & 72.3          & 62.5          & 59.3          \\
\textbf{TokenCarve (random, F1)}  & -             & \textbf{87}   & \textbf{86.6} & \textbf{84.9} & \textbf{85.3} & \textbf{85}   & \textbf{81.7} & \textbf{80}   \\ \hline
\end{tabular}}
\caption{Result on POPE}
\label{tab:Result on POPE}
\vspace{-3mm}
\end{table}

\begin{table}[h!]
\resizebox{\columnwidth}{!}{%
\fontsize{7pt}{10}\selectfont
\begin{tabular}{c|cccccccc}
\hline
\textbf{Method/Visual token}   & \textbf{576}  & \textbf{333}  & \textbf{288}  & \textbf{192}  & \textbf{144}  & \textbf{128} & \textbf{72}   & \textbf{64}   \\ \hline
\textbf{Fastv}      & \textbf{69.5} & 68.9          & 68.9          & 68.9          & \textbf{68.9} & \textbf{69}  & \textbf{69.5} & \textbf{69.9} \\
\textbf{TokenCarve} & -             & \textbf{69.2} & \textbf{69.3} & \textbf{69.1} & 68.8          & 68.9         & 69            & 69.7          \\ \hline
\end{tabular}}
\caption{Result on SQA}
\label{tab:Result on SQAIMG}
\vspace{-3mm}
\end{table}

\begin{table}[h!]
\resizebox{\columnwidth}{!}{%
\fontsize{7pt}{10}\selectfont
\begin{tabular}{c|cccccccc}
\hline
\textbf{Method/Visual token} & \textbf{576} & \textbf{333}  & \textbf{288}  & \textbf{192}  & \textbf{144}  & \textbf{128}  & \textbf{72}   & \textbf{64}   \\ \hline
\textbf{Fastv}               & \textbf{50}  & \textbf{50.4} & \textbf{50.5} & \textbf{50.9} & \textbf{51.4} & \textbf{51.3} & \textbf{51.7} & \textbf{51.8} \\
\textbf{TokenCarve}          & -            & 50.4          & 50.7          & 50.9          & 50.9          & 51            & 51.5          & 51.4          \\ \hline
\end{tabular}
}
\caption{Result on VizWizVQA}
\label{tab:Result on VizWizVQA}
\vspace{-3mm}
\end{table}

\begin{table}[h!]
\resizebox{\columnwidth}{!}{%
\fontsize{7pt}{10}\selectfont
\begin{tabular}{c|cccccccc}
\hline
\textbf{Method/Visual token}    & \textbf{576} & \textbf{333} & \textbf{288} & \textbf{192} & \textbf{144} & \textbf{128} & \textbf{72} & \textbf{64} \\ \hline
\textbf{Fastv}      & \textbf{58.2}         & \textbf{58.3}         & \textbf{58.2}         & 58           & 57.4         & 57           & 55.6        & 55.1        \\
\textbf{TokenCarve} & -            & 58.2         & 58           & \textbf{58.4}         & \textbf{58.3}         & \textbf{58.1}         & \textbf{57.2}        & \textbf{57}          \\ \hline
\end{tabular}}
\caption{Result on TextVQA}
\label{tab:Result on TextVQA}
\vspace{-3mm}
\end{table}

\begin{table}[h!]
\resizebox{\columnwidth}{!}{%
\fontsize{7pt}{10}\selectfont
\begin{tabular}{c|cccccccc}
\hline
\textbf{Method/Visual token}      & \textbf{576}  & \textbf{333}  & \textbf{288}  & \textbf{192} & \textbf{144}  & \textbf{128}  & \textbf{72}   & \textbf{64}   \\ \hline
\textbf{Fastv}      & \textbf{78.5} & 78            & 77.7          & 75.9         & 74            & 73.2          & 67.7          & 66.3          \\
\textbf{TokenCarve} & -             & \textbf{78.4} & \textbf{78.4} & \textbf{78}  & \textbf{77.5} & \textbf{77.3} & \textbf{75.4} & \textbf{74.8} \\ \hline
\end{tabular}}
\caption{Result on VQAV2}
\label{tab:Result on VQAV2}
\vspace{-3mm}
\end{table}

\begin{table}[h!]
\resizebox{\columnwidth}{!}{%
\fontsize{7pt}{10}\selectfont
\begin{tabular}{c|cccccccc}
\hline
\textbf{Method/Visual token}     & \textbf{576} & \textbf{333} & \textbf{288} & \textbf{192} & \textbf{144} & \textbf{128} & \textbf{72} & \textbf{64} \\ \hline
\textbf{Fastv}      & \textbf{31}           & \textbf{31.8}         & 31.1         & 28.9         & 27.9         & 26.7         & 27.6        & 26.1        \\
\textbf{TokenCarve} & -            & 31.1         & \textbf{31.8}         & \textbf{30.4}         & \textbf{30}           & \textbf{29.5}         & \textbf{28.5 }       & \textbf{29.3}        \\ \hline
\end{tabular}}
\caption{Result on MM-Vet}
\label{tab:Result on MM-Vet}
\vspace{-3mm}
\end{table}

\begin{table}[h!]
\resizebox{\columnwidth}{!}{%
\fontsize{7pt}{10}\selectfont
\begin{tabular}{c|cccccccc}
\hline
\textbf{Method/Visual token}     & \textbf{576} & \textbf{333} & \textbf{288} & \textbf{192} & \textbf{144} & \textbf{128} & \textbf{72} & \textbf{64} \\ \hline
\textbf{Fastv}      & \textbf{74.6}         & 72.5         & 72.7         & 72           & 70.6         & 69.6         & 60          & 56.5        \\
\textbf{TokenCarve} & -            & \textbf{74.1}         & \textbf{74.2}         & \textbf{73.1}         & \textbf{71.7}         & \textbf{71.3}         & \textbf{64.6}        & \textbf{63.2}        \\ \hline
\end{tabular}
}
\caption{Result on OCRVQA}
\label{tab:Result on OCRVQA}
\vspace{-3mm}
\end{table}

\begin{table}[h!]
\resizebox{\columnwidth}{!}{%
\fontsize{10pt}{12}\selectfont
\begin{tabular}{c|cccccccccc}
\hline
\textbf{Method/Visual token} & \textbf{576}  & \textbf{288}  & \textbf{155}  & \textbf{144}  & \textbf{120}  & \textbf{90 }  & \textbf{72}   & \textbf{60}   & \textbf{50}   & \textbf{40}   \\ \hline
\textbf{Fastv}               & \textbf{74.8} & 75.3          & 74.9 & 74.2 & 73.2 & 72.7 & 71.5 & 69.6 & 68.2 & 65.9 \\
\textbf{TokenCarve}          & -              & \textbf{75.9}  & \textbf{75.5} & \textbf{75.5}& \textbf{74.4} & \textbf{73.1} & \textbf{72.1} & \textbf{70.9} & \textbf{70.1} & \textbf{68.1} \\ \hline
\end{tabular}
}
\caption{Result on AOKVQA}
\label{tab:Result on AOKVQA}
\vspace{-3mm}
\end{table}

\newpage
\subsection{Results on LLaVA-13B}
We complement the experimental results of LLaVA-13B on TextVQA, MM-Vet and POPE. On all three datasets we achieve SOTA results. On the POPE dataset, when the number of visual tokens is compressed to 64, TokenCarve still maintains a performance of 83.7 while FastV drops to 69.2. This illustrates that our compression strategy of minimal information loss preserves the vast majority of information in visual tokens.

\begin{table}[h!]
\resizebox{\columnwidth}{!}{%
\fontsize{7pt}{10}\selectfont
\begin{tabular}{c|cccccccc}
\hline
\textbf{Method/Visual token}          & \textbf{576}  & \textbf{333}  & \textbf{288}  & \textbf{192}  & \textbf{144}  & \textbf{128}  & \textbf{72}   & \textbf{64}   \\ \hline
\textbf{Fastv (13B)}      & \textbf{61.2} & 61            & 60.9          & 60.5          & 60.1          & 59.7          & 56.6          & 56.1          \\
\textbf{TokenCarve (13B)} & -             & \textbf{61.5} & \textbf{60.9} & \textbf{60.8} & \textbf{60.8} & \textbf{60.6} & \textbf{59.6} & \textbf{59.3} \\ \hline
\end{tabular}}
\caption{Result on TextVQA (13B)}
\label{tab:Result on TextVQA 13B}
\vspace{-3mm}
\end{table}

\begin{table}[h!]
\resizebox{\columnwidth}{!}{%
\fontsize{7pt}{10}\selectfont
\begin{tabular}{c|cccccccc}
\hline
\textbf{Method/Visual token}           & \textbf{576}  & \textbf{333}  & \textbf{288}  & \textbf{192} & \textbf{144}  & \textbf{128}  & \textbf{72}   & \textbf{64}   \\ \hline
\textbf{Fastv (13B)}      & \textbf{36.1} & 34.7          & 35.2          & \textbf{35}  & 32.9          & 34            & 33.8          & 31.2          \\
\textbf{TokenCarve (13B)} & -             & \textbf{36.1} & \textbf{35.3} & 34.4         & \textbf{35.6} & \textbf{35.1} & \textbf{34.5} & \textbf{33.9} \\ \hline
\end{tabular}}
\caption{Result on MM-Vet (13B)}
\label{tab:Result on MM-Vet 13B}
\vspace{-3mm}
\end{table}

\begin{table}[h!]
\resizebox{\columnwidth}{!}{%
\fontsize{7pt}{10}\selectfont
\begin{tabular}{c|cccccccc}
\hline
\textbf{Method/Visual token}             & \textbf{576}  & \textbf{333}  & \textbf{288}  & \textbf{192}  & \textbf{144} & \textbf{128}  & \textbf{72}   & \textbf{64}   \\ \hline
\textbf{Fastv (13B)}      & \textbf{86.4} & 85.8          & 85.6          & 83.3          & 79.8         & 78            & 70            & 69.2          \\
\textbf{TokenCarve (13B)} & -             & \textbf{86.6} & \textbf{86.8} & \textbf{86.8} & \textbf{86}  & \textbf{85.8} & \textbf{84.1} & \textbf{83.7} \\ \hline
\end{tabular}}
\caption{Result on POPE (13B)}
\label{tab:Result on POPE 13B}
\vspace{-3mm}
\end{table}


\end{document}